\documentclass[runningheads]{llncs}
\usepackage[T1]{fontenc}
\usepackage{subcaption}
\usepackage{float}
\usepackage{graphicx}
\usepackage{multirow}
\usepackage{array}
\usepackage{amsmath}
\newcommand{\dvpsnr}{$DV_{psnr}$}
\newcommand{\dvloss}{$DV_{loss}$}
\begin{document}
\title{Image valuation in NeRF-based 3D reconstruction}
%
%
%
\author{Grigorios Aris Cheimariotis\orcidID{0000-0003-1984-5030} \and
Antonis Karakottas\orcidID{0000-0003-0328-8671} \and
Vangelis Chatzis\orcidID{0000-0002-6717-5185} \and
Angelos Kanlis\orcidID{0009-0005-0836-9325} \and
Dimitrios Zarpalas\orcidID{0000-0002-9649-9306}}
\authorrunning{G-A. Cheimariotis et al.}
%
\institute{
Information Technologies Institute (ITI),Centre for Research and Technology Hellas (CERTH),Thessaloniki, Greece\\
\email{\{acheimar, ankarako, chatzise, a.kanlis, zarpalas\}@iti.gr}\\ 
}
\maketitle              
\begin{abstract}

Data valuation and monetization are becoming increasingly important across domains such as eXtended Reality (XR) and digital media. In the context of 3D scene reconstruction from a set of images -whether casually or professionally captured - not all inputs contribute equally to the final output. Neural Radiance Fields (NeRFs) enable photorealistic 3D reconstruction of scenes by optimizing a volumetric radiance field given a set of images. However, in-the-wild scenes often include image captures of varying quality, occlusions, and transient objects, resulting in uneven utility across inputs. In this paper we propose a method to quantify the individual contribution of each image to NeRF-based reconstructions of in-the-wild image sets. Contribution is assessed through reconstruction quality metrics based on PSNR and MSE. We validate our approach by removing low-contributing images during training and measuring the resulting impact on reconstruction fidelity.

\end{abstract}

\newcommand\blfootnote[1]{%
  \begingroup
  \renewcommand\thefootnote{}\footnote{#1}%
  \addtocounter{footnote}{-1}%
  \endgroup
}

\blfootnote{© Springer-Verlag GmbH Germany, part of Springer Nature 2025. This is the author’s accepted manuscript (preprint) of an article published in Castrillón-Santana, M., et al. Computer Analysis of Images and Patterns. CAIP 2025. Lecture Notes in Computer Science, vol 15621. Springer, Cham. The final authenticated version is available online at: 10.1007/978-3-032-04968-1-32. Personal use of this material is permitted. Permission from Springer must be obtained for all other uses, including reproduction, distribution, or reuse in other works. This preprint has not undergone any post-submission improvements or corrections. This work was supported by the European Union’s Horizon Europe programme under grant number 101070250 “XReco” (https://xreco.eu/). The computational resources were granted with the support of GRNET.}

\vspace{-1.5em}
\section{Introduction}
3D scene reconstruction has broad applications, including virtual tourism \cite{helmy2024navigating}, cultural heritage preservation \cite{mazzacca2023nerf}, or healthcare \cite{molaei2023implicit}. Recent advances enable photorealistic novel view synthesis from 2D image collections. Among the most prominent methods are Neural Radiance Fields (NeRFs) \cite{mildenhall2021nerf} and 3D Gaussian Splatting \cite{kerbl20233d}, both of which learn scene representations directly from images. The Phototourism dataset \cite{jin2021image} offers a compelling use case, enabling the reconstruction of explorable 3D scenes from in-the-wild photo collections. However, it also introduces challenges such as variable image quality, transient objects, and appearance inconsistencies. Techniques like NeRF-W \cite{martin2021nerf} and HA-NeRF \cite{chen2022hallucinated} have demonstrated effectiveness in addressing these issues.

However, an image with transient objects, differences in appearance and/or low quality can negatively impact reconstruction quality and it is preferable to be excluded from training.  In NeRF-W, images undergo a coarse pre-filtering step based on two heuristic criteria. Firstly, using Neural Image Assessment (NIMA)\cite{talebi2018nima} —a model producing continuous scores (1 to 10) based on human-rated aesthetic judgments—images with a NIMA score below 3 are omitted. Secondly, images are filtered out where transient objects, detected by a DeepLab v3 \cite{chenrethinking} model, occupy more than 80\% of the image's area. These methods result in a binary classification (keep/discard) applied prior to NeRF training, without explicit validation of an image's actual impact on the NeRF's reconstruction performance during training. Therefore, it cannot measure the potential impact of a training image in the reconstructed scene. Therefore, in NeRF-W, images are heuristically classified as either helpful or harmful; however, this categorization is not explicitly validated - images deemed harmful may still contribute positively to training, and vice versa.

This study aims to assign continuous contribution scores to individual images based on their impact on the final NeRF-based scene reconstruction. The motivation stems from two key data valuation objectives: (1) enabling fair compensation to different data providers in data marketplaces, and (2) improving the performance of data-driven systems. Reconstruction accuracy is evaluated using Peak Signal-to-Noise Ratio (PSNR) computed on a set of held-out test images for each scene.

Numerous studies have explored data valuation methods in the context of machine and deep learning, reflecting growing interest in the field \cite{ghorbani2019data,guo2020fastif,yoon2020data}. In this work, we draw inspiration from Data Valuation using Reinforcement Learning (DVRL) \cite{yoon2020data}, particularly its ability to estimate sample importance without retraining the model - an appealing property given the computational cost of NeRF training. Unlike, DVRL, which uses a learned policy to adjust training dynamics, our approach repurposes the reward signal directly as a continuous contribution score for each image in the NeRF pipeline.

Rather than proposing a new fundamental data valuation method, this work explores the applicability and adaptation of existing data valuation principles (specifically DVRL) to the unique demands of NeRF training and 3D rendering. Our main contributions lie in this adaptation, the comprehensive examination of various metrics, and the evaluation of their robustness and consistency. Additionally, we achieve the efficient extraction of explicit 'contribution' scores that are fast to obtain (especially compared to fundamental data valuation methods applied to NeRFs) and whose empirical utility is demonstrated by our results.

\section{Related Work}
\subsection{Neural Radiance Fields in the Wild}
Neural Radiance Fields (NeRFs) \cite{mildenhall2021nerf} have become a powerful approach for synthesizing photorealistic views of scenes from sparse input images. By modelling a scene as a continuous 5D function parametrized by a neural network, NeRFs deliver impressive results in controlled environments. However, their performance often deteriorates in real-world scenarios due to challenges such as transient occluders, variable lighting, weather conditions, and scene complexity. Several variants have been developed to address these issues. Ha-NeRF \cite{chen2022hallucinated} improves scene representation through enhanced hierarchical sampling, while NeRF-W \cite{martin2021nerf} introduces a mechanism to separate static and transient components, allowing more robust reconstructions from in-the-wild-photo collections. Additionally, NeRF On-the-go \cite{ren2024nerf} leverages pre-trained DINOv2 \cite{oquab2023dinov2} features to predict per-pixel uncertainty and remove transient objects from neighboring rays via cosine similarity in feature space.

\subsection{Data valuation}
Traditional image data valuation methods include Data Shapley \cite{ghorbani2019data}, which computes the marginal contribution of each through game-theoretic Shapley values but is computationally prohibitive for large datasets, influence functions, which estimate the effect of removing an image on a model’s loss but rely on unstable Hessian–gradient approximations and local convexity assumptions. These approaches suffer from high computational overhead, limited scalability, and static valuation that cannot adapt during training. In contrast, Data Valuation using Reinforcement Learning (DVRL) frames valuation as a sequential decision process: a neural data-value estimator learns sampling probabilities via policy gradients, guided by rewards from a small validation set. DVRL avoids costly second-order calculations, scales efficiently through multinomial sampling, and adaptively updates values online to reflect evolving model and data characteristics.

For Neural Radiance Fields (NeRFs), which rely on multi-view image consistency to reconstruct 3D scenes, the quality and relevance of each image play a crucial role in performance. In real-world scenes, transient objects such as people, vehicles, or lighting variations can degrade the reconstruction quality. Recent approaches, such as NeRF-W, incorporate image valuation mechanisms to mitigate this by leveraging a learned weighting strategy during training.  Complementary techniques include the use of pre-trained Neural Image Assessment (NIMA) models to predict aesthetic or quality-related properties, helping identify and exclude images with low utility, particularly those containing significant transient content. This combination of learned valuation and perceptual quality assessment enables more robust training pipelines, especially for unconstrained datasets, such as Phototourism \cite{jin2021image}.

\section{Method}
\begin{figure}[t]
    \includegraphics[width=\textwidth]{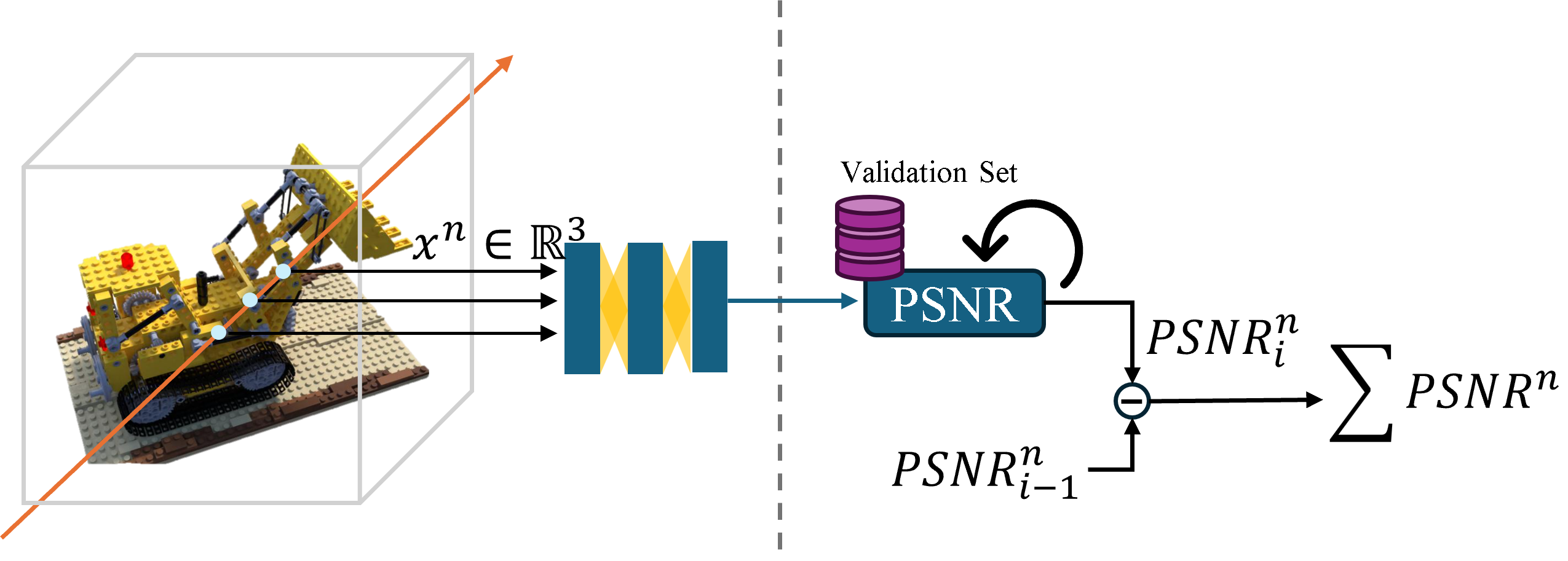}
    \caption{Method overview: During NeRF training, we evaluate a small validation set after each training iteration $i$, recording the PSNR associated with image $I^n$. When image $I^n$ is revisited, we compute the change in PSNR by subtracting the previous value from the current one. These PSNR differences are then aggregated to estimate each image's \dvpsnr{}.
    }
    \label{fig:overview}
\end{figure}

The proposed valuation scheme is compatible with any NeRF variant that samples rays from a single image per training iteration. At each iteration, we evaluate the PSNR on a fixed test set and associate it with the image sampled at the moment $I^n$. For subsequent appearances of $I^n$, we compute the change in PSNR relative to its previous PSNR value. These PSNR deltas are then aggregated across iterations to quantify each image's overall influence on reconstruction quality, which we refer to as \dvpsnr{} (Figure \ref{fig:overview}). In the following sections we first  describe our training setup (Section \ref{ssec:train-config}). In Section \ref{ssec:contri-scores} we describe our approach to identify fair metrics for valuating each image's contribution score. In Section \ref{ssec:calc-scores} we present the metric utilized and, finally, in Section \ref{sec:exps} we present our experimental setup and results.

\subsection{Training configuration}
\label{ssec:train-config}
Our NeRF training set-up follows a conventional structure. Each training step samples rays from a single image. All training images are used once per epoch, and the order of image sampling is shuffled at the beginning of each epoch. This configuration is consistent with standard NeRF training practices and allows for systematic tracking of each image's influence on validation set performance. Impact is measured as the change in test $L1$ loss and PSNR.
This design also offers two key advantages for image valuation: first, the per-image sampling enables attribution of performance changes to specific inputs; second, the randomized order across epochs helps mitigate ordering bias and supports robust aggregation of impact scores. Together, these factors make the training process suitable for analysing contribution of data in dynamic, real-world datasets.

\subsection{Identifying fair contribution scores}
\label{ssec:contri-scores}
Since the model's parameters evolve throughout training, the impact of a given image depends on the model's current state. To ensure a fair comparison across images, we consider three complementary strategies: (a) Weighting the contribution of each image by training progress, for instance, assigning higher importance to updates made in later training stages. (b) Aggregating impact estimates across epochs, to account for randomness in image sampling order. (c) Reverting the model to its pre-update state before measuring the effect of each training image.

Option (a), while intuitively appealing, was not adopted in this study due to the lack of a principled framework to define the weighting function. Any choice of weight over training iterations, such as linear, exponential, or heuristic decay, would introduce arbitrary assumptions that are not grounded in theoretical or empirical validation. Moreover, weighting alone does not fully account for the interaction between the image and the model state, which is itself non-linearly dependent on training history.

Option (b) was selected as the main approach due to its practicality and empirical stability. It involves aggregating the impact of each image across all epochs, excluding the first epoch to avoid the outsized influence of early updates when the model is not trained. The resulting metrics - \dvpsnr{}   and \dvloss{} - are computed as the cumulative change in validation PSNR and L1 loss, respectively, each time an image is used. Although random image ordering across epochs does not fully eliminate model-state bias, we observed that the model quickly converges to a stable performance baseline (e.g, validation PSNR $\approx$ 14) after the first epoch. This suggests that, in later epochs, each image is processed under a comparably trained model, mitigating unfairness introduced by early training dynamics.

Option (c) was considered as a more controlled alternative, where the model would be reverted to a consistent state before each image-specific update. However, this approach was found to be computationally intensive and unstable in practice. In particular, measurements taken in the final epoch under this scheme exhibited high variance across runs with different random seeds, indicating that local updates can lead to unpredictable validation effects. This instability undermines the reliability of  impact estimates and limits their utility for consistent image valuation.

To assess the reproducibility of image contribution scores, we identified and measured key source of variability - namely, the random ordering of training images and the random sampling of pixels within image. These are influenced by the random seed used during training. Correlation analyses across runs with different seeds revealed that reproducibility improves with more epochs and larger per-image sampling, but this trend held consistently only for one metric. Notably, the aggregated PSNR difference (\dvpsnr{}) demonstrated a satisfactory correlation when 500 pixel-rays per image were used in each iteration.

\subsection{Contribution scores}
\label{ssec:calc-scores}
At each training step, the L1 loss and PSNR were computed on a fixed validation set to calculate contribution scores for individual images. The L1 loss is a commonly used loss function in regression tasks and also drives NeRF training. It measures the absolute difference between the predicted values and the ground truth values.

\noindent
\begin{equation}
\mathcal{L}_{L1} = \sum_{i=1}^{N} | I_i - \hat{I}_i |
\end{equation}
where $N$ is the number of data points, $I_i$ is the original image, and $\hat{I}_i$ is the NeRF rendered one. L1 loss is considered more robust to outliers than Mean Squared Error (MSE), as it penalizes deviations linearly rather than quadratically, thereby limiting the influence of large errors.

PSNR is calculated using the MSE as follows:

\noindent
\begin{equation}
\text{MSE} = \frac{1}{MN} \sum_{i=1}^{M} \sum_{j=1}^{N} \left( I_{i,j} - \hat{I}_{i,j} \right)^2
\end{equation}
where $I_{i, j}$ is the original image, $\hat{I}_{i,j}$ is the reconstructed image, and $(M, N)$ are the dimensions of the image.
The PSNR in decibels is then given by:

\noindent
\begin{equation}
\text{PSNR} = 10 \cdot \log_{10} \left( \frac{L^2}{\text{MSE}} \right)
\end{equation}
where  $L$  is the maximum possible pixel value of the image (e.g., 255 for 8-bit unsigned integer images).

\subsubsection{PSNR Change and Aggregation for a Training Image}

The instantaneous PSNR difference in our experiments was calculated with:

\noindent
\begin{equation}
\Delta \text{PSNR}_i^{(t)} = \text{PSNR}(I_i;\theta^{(t+1)}) - \text{PSNR}(I_i;\theta^{(t)})
\end{equation}
where $\text{PSNR}^{(t)}$ is the calculated PSNR over the validation set at training step $t$. $I_i$ is the $i$-th training image, $\theta^{(t)}$ are the NeRF MLP parameters before training on $I_i$, $\theta^{(t+1)}$ are the NeRF MLP parameters after training on $I_i$, and $\Delta \text{PSNR}_i^{(t)}$ is the change in PSNR due to training on $I_i$ at step $t$. This quantifies the immediate effect of the training image \( I_i \) on test PSNR. The $\Delta \text{PSNR}_i^{(t)}$  values were aggregated for each image for all epochs, excluding the first. The outcome contribution score is referred to as \dvpsnr{}.

\section{Experiments}
\label{sec:exps}

\subsubsection{Dataset}
We evaluated our method using 4 scenes from the PhotoTourism dataset \cite{jin2021image}, which contains internet-sourced photo collections of famous landmarks with significant viewpoint, transiency and appearance variations. These scenes present challenges such as inconsistent lighting, occlusions from crowds, and diverse camera parameters, making them ideal benchmarks for assessing robustness in unconstrained, in-the-wild conditions. 

We focused on the Brandenburg Gate scene to evaluate the consistency of the proposed contribution score. This scene consists of 1363 images. NeRF-W \cite{martin2021nerf} reserved 10 for testing, 763 for training, 96 for validation, and the rest are excluded from the process due to the automatic assessment which they proposed \cite{talebi2018nima}. In addition, we also measured contribution scores for three more scenes (Sacre Coeur, Taj Mahal, Trevi Fountain) to assess the impact of training NeRF with images highly valued by our proposed method.

\subsubsection{Reproducibility evaluation with correlation}
\begin{figure}[t!]
\centering
\begin{subfigure}[b]{0.45\textwidth}
    \centering
    \includegraphics[width=0.8\linewidth]{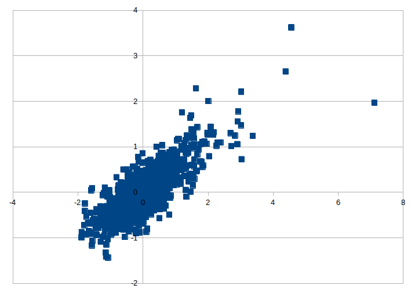} 
    \caption{Aggregated PSNR difference}
    \label{fig:psnrcorr}
\end{subfigure}
\hfill
\begin{subfigure}[b]{0.45\textwidth}
    \centering
    \includegraphics[width=0.8\linewidth]{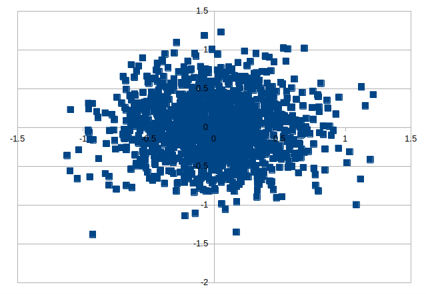} 
    \caption{Aggregated L1 loss difference}
    \label{fig:l1corr}
\end{subfigure}

\vspace{1em} 

\begin{subfigure}[b]{0.45\textwidth}
    \centering
    \includegraphics[width=0.8\linewidth]{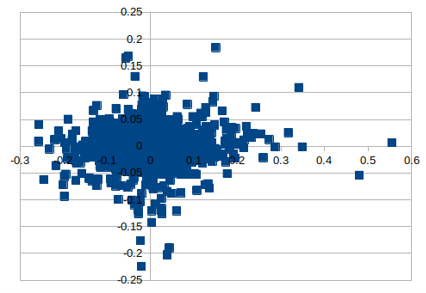} 
    \caption{Last epoch PSNR difference}
    \label{fig:lastpsnr}
\end{subfigure}
\hfill
\begin{subfigure}[b]{0.45\textwidth}
    \centering
    \includegraphics[width=0.8\linewidth]{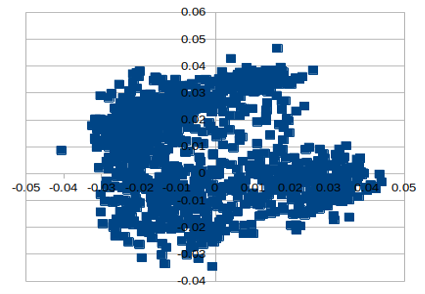} 
    \caption{PSNR difference of each image from the same trained state}
    \label{fig:psnrtrained}
\end{subfigure}

\caption{Contribution score correlations on the Brandenburg dataset. The x-axis shows contribution scores from one training run, and the y-axis from a different run with a different seed. Each (x, y) point compares scores for the same image.}
\label{fig:scoresandCorrelation}
\end{figure}

A primary objective of this study was to develop contribution metrics that are minimally affected by the image loading sequence. As shown in Figure \ref{fig:scoresandCorrelation}, the \dvpsnr{} scores exhibit strong consistency across runs, achieving a correlation coefficient of 0.8 for the Brandenburg scene.
In each training step, PSNR was measured in a small validation set (10-14 images). For the Brandenburg scene, all images except the test images and validation images were valuated, totalling 1310 images. For reproducibility evaluation, for the other three scenes, a subset of 100 images was valuated twice with different seeds to measure consistency, and the correlation coefficients were 0.90 (Sacre Coeur), 0.87 (Trevi Fountain), and 0.86 (Taj Mahal). However, to measure the impact of the training composition, all images except the test and validation images, were valuated once for all the scenes.  

\subsubsection{Impact of Training Set Composition}
Another primary objective was to assess the impact of using different training subsets. To expedite the computation of the contribution scores, PSNR was measured at each training iteration. For the Brandenburg scene, based on the calculated scores, 716 images were identified as having a positive impact on improving PSNR in the small validation set. To evaluate the effect of these selections, two different training sets were compared using 43 held-out test images, which were not part of the training or the validation sets used for computing contribution scores. The training set selected by the proposed method achieved higher PSNR in a  10-image validation set but lower in a larger 43-image test set that included finer details, as demonstrated in Table \ref{tab:scores}. 

\begin{table}[tbp]
    \centering
    \caption{PSNR achieved for the same pipeline with NeRF-W training set and with  training set which was selected by the proposed data valuation. }
    \label{tab:scores}

        \vspace{2mm}

    \begin{tabular}{|l|c|c|c|c|c|c|c|c|}
        \hline
        \multirow{2}{*}{PSNR} & \multicolumn{2}{|c|}{Brandenburg } & \multicolumn{2}{|c|}{ Sacre Coeur} & \multicolumn{2}{|c|}{Taj Mahal} & \multicolumn{2}{|c|}{Trevi Fountain} \\
        \cline{2-9} 
        \cline{2-9} %
        & val  & test  & val  & test  & val  & test & val  & test\\
        \hline
        NeRF-W  & 19.27 & \textbf{17.72} &  16.94 & 15.62  &  17.93 &  16.22 &  17.51 & 17.33\\
        \hline
        \dvpsnr{}  & \textbf{19.96} & 16.79 & \textbf{17.45} & \textbf{16.23} &  \textbf{18.33} &  \textbf{16.55} &  \textbf{17.65} &  \textbf{17.50}\\
        \hline
    \end{tabular}
\end{table}

Also, in Table \ref{tab:scores}, it is observed that, in the other 3 scenes, the training sets selected by our data valuation framework consistently result in higher PSNR on the test sets compared to the training set selected in NeRF-W. In these 3 scenes, validation and test sets are of almost equal size (Sacre Coeur 11:11, Taj Mahal 14:13, Trevi Fountain 10:9)  and present a more balanced and representative distribution of images.

\begin{figure}[t!]
\centering
\begin{subfigure}[b]{0.22\textwidth}
    \includegraphics[width=\textwidth]{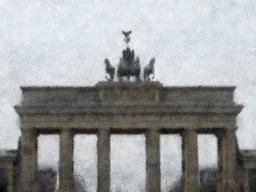}
    \caption{Test image rendered by NeRF trained with 716 images selected by \dvpsnr{}.}
\end{subfigure}%
\hspace{0.5 em}%
\begin{subfigure}[b]{0.22\textwidth}
    \centering
    \includegraphics[width=\textwidth]{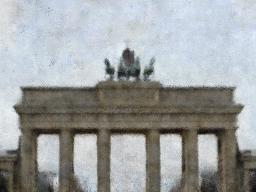}
    \caption{Test image rendered by NeRF trained with 763 images (NeRF-W selection).}
\end{subfigure}
\begin{subfigure}[b]{0.22\textwidth}
    \includegraphics[width=\textwidth]{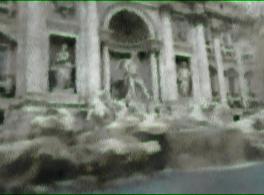}
    \caption{Test image rendered by NeRF trained with 1717 images selected by \dvpsnr{}.}
\end{subfigure}%
\hspace{0.5 em}
\begin{subfigure}[b]{0.22\textwidth}
    \centering
    \includegraphics[width=\textwidth]{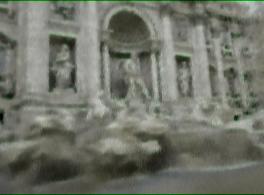}
    \caption{Test image rendered by NeRF trained with 1716 images(NeRF-W selection).}
\end{subfigure}
\begin{subfigure}[b]{0.22\textwidth}
    \includegraphics[width=\textwidth]{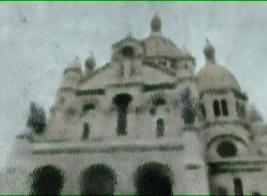}
    \caption{Test image rendered by NeRF trained with 954 images selected by \dvpsnr{}.}
\end{subfigure}%
\hspace{0.5 em}
\begin{subfigure}[b]{0.22\textwidth}
    \centering
    \includegraphics[width=\textwidth]{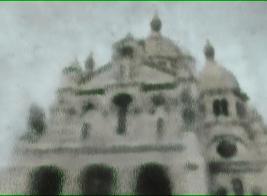}
    \caption{Test image rendered by NeRF trained with 950 images(NeRF-W selection).}
\end{subfigure}
\begin{subfigure}[b]{0.22\textwidth}
    \includegraphics[width=\textwidth,height=57pt]{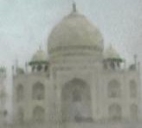}
    \caption{Test image rendered by NeRF trained with 899 images selected by \dvpsnr{}.}
\end{subfigure}%
\hspace{0.5 em}
\begin{subfigure}[b]{0.22\textwidth}
    \centering
    \includegraphics[width=\textwidth,height=57pt]{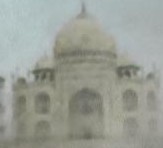}
    \caption{Test image rendered by NeRF trained with 896 images(NeRF-W selection).}
\end{subfigure}
\caption{Rendered images from NeRF trained on \dvpsnr{} selections (a),(c),(e),(g) and on NeRF-W training sets (b),(d),(f),(h). PSNR = (a) 19.51, (b)  17.59, (c)  21.3, (d)  20.7,(e) 17.31, (f)  17.03, (g)  24.38, (h)  22.9.}
\label{fig:resultimages}
\end{figure}

In addition, for test views taken from more distant vantage points, the \dvpsnr{}-selected training set yields better performance, as illustrated in Figure \ref{fig:resultimages} (a-b). The model trained on the proposed dataset more accurately reconstructs the statue on top of the Brandenburg Gate and achieves a higher PSNR ($19.51$). In the example test images of the Trevi Fountain scene and Sacre Coeur scene (Figure \ref{fig:resultimages} (c-f)), it is noticed that some details mainly on the bottom side and on the left side are sharper in the images rendered by the NeRF trained on  \dvpsnr{} selection. In the example test image of Taj Mahal (Figure \ref{fig:resultimages} (g-h)), there are no obvious differences. However, the image rendered by the NeRF trained on \dvpsnr{} selection achieved higher PSNR. Although the data valuation framework may not be fully optimal, it demonstrates a reasonable ability to discriminate between useful and less useful training images.

\begin{figure}[t!]
\centering
\begin{subfigure}[b]{0.45\textwidth}
    \includegraphics[width=\textwidth]{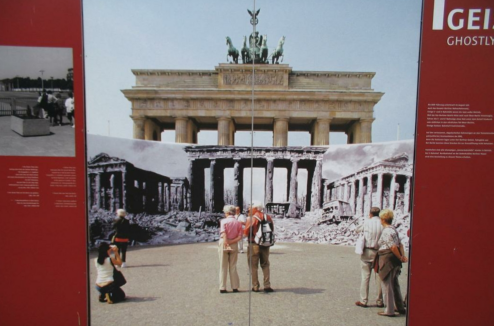}
    \caption{Image included in the NeRF-W training set, is valuated as "harmful" for training NeRF by \dvpsnr{}} 
\end{subfigure}%
\hspace{0.2em}%
\begin{subfigure}[b]{0.45\textwidth}
    \centering
    \includegraphics[width=\textwidth]{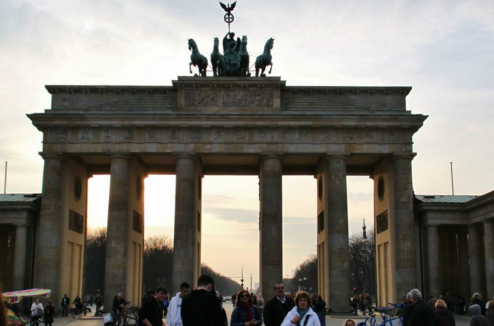}
    \caption{Image excluded from NeRF-W training-scored a high contribution \dvpsnr{} score.}
\end{subfigure}
\caption{NeRF-W NIMA valuation and \dvpsnr{} valuation.}
\label{fig:valuatedimages}
\end{figure}

As shown in Figure \ref{fig:valuatedimages}, the proposed data valuation framework identified an image included in NeRF-W's training set - although it contains numerous transient objects - as potentially harmful. In contrast, an image excluded from the NeRF-W training, validation, and test sets was rated highly valuable by our method. Each image presents both strengths and limitations: image (a) is brighter and may reveal more fine detail in unoccluded regions, while image (b) is darker with some less distinct areas but exhibits significantly fewer occlusions. These differences in valuation may also be influenced by the image downscaling used during analysis.

\section{Discussion}
In this paper, we present a method for image valuation in NeRF-based 3D reconstruction. The proposed approach yields meaningful results, demonstrated both quantitatively and qualitatively, by showing that training on a subset of positively scored images can improve reconstruction performance in certain cases.

A primary focus of the study was the reproducibility of the contribution scores. We evaluated scores based on PSNR for their consistency across runs with different random seeds. Only aggregated PSNR-based scores (\dvpsnr{}) exhibited a strong correlation between training runs.

Crucially, the practical value of the contribution scores was assessed by their effectiveness in guiding the selection of training sets. The \dvpsnr{}-selected dataset slightly outperformed the NeRF-W \cite{martin2021nerf} training set and achieved better results for distant viewpoints. This is due to image downscaling during valuation. It is arguable that high-performing NeRF would deal differently with full-resolution images. However, scores derived from downscaling were shown to be meaningful at double the valuation resolution, producing consistent PSNR improvements (Table \ref{tab:scores}). From this, we could hypothesize that data valuation scores may be transferable to other resolutions.

One limitation of this study is its reliance on two versions of an experimental NeRF pipeline, which do not reach state-of-the-art performance. For efficiency, we used NeRFs inspired by HA-NeRF \cite{chen2022hallucinated} and NeRF-W \cite{martin2021nerf}, which reduce training time, but struggle with accuracy. However, the proposed valuation method is designed to be model-agnostic and could be integrated into any NeRF architecture. If the constraint of using only one image in each step of contribution scores' extraction is met, the method independently computes explicit image contribution scores.

Due to memory and time constraints, incorporating data valuation into high-end pipelines remains a challenge, as contribution scoring adds computational overhead. On an NVIDIA GeForce RTX 3060 (12GB), a standard training step takes 0.185 seconds, with \dvpsnr{}'s contribution scoring adding approximately 0.5 seconds per step. For a typical 50,000-iteration run, this is about 9.5 hours of total training time. Compared to data valuation methods like Data Shapley, which demand multiple re-runs, our approach offers a considerably faster alternative. Although these timings are hardware-dependent and can be faster with more powerful resources, we believe this is a manageable cost for the continuous, explicit image valuation provided. Our future research aims to further optimize this by directly inferring contribution scores from a learnable model.

This work focused exclusively on the PhotoTourism dataset \cite{jin2021image}, given its rich variation in image quality and its frequent use in NeRF evaluation. Future work should explore the application of \dvpsnr{} to other datasets, NeRF variants, as well as more recent 3D Gaussian Splatting techniques such as \cite{kulhanek2024wildgaussians}, to further validate its generalizability.

\begin{credits}
\subsubsection{\ackname} This work was supported by the European Union’s Horizon Europe programme under grant number 101070250  «XRECO: XR media eCOsystem»
(https://xreco.eu/). The computational
resources were granted with the support of GRNET.
\end{credits}

\bibliographystyle{splncs04}
\bibliography{bibliography}

@article{mildenhall2021nerf,
  title={Nerf: Representing scenes as neural radiance fields for view synthesis},
  author={Mildenhall, Ben and Srinivasan, Pratul P and Tancik, Matthew and Barron, Jonathan T and Ramamoorthi, Ravi and Ng, Ren},
  journal={Communications of the ACM},
  volume={65},
  number={1},
  pages={99--106},
  year={2021},
  publisher={ACM New York, NY, USA}
}

@inproceedings{yoon2020data,
  title={Data valuation using reinforcement learning},
  author={Yoon, Jinsung and Arik, Sercan and Pfister, Tomas},
  booktitle={International Conference on Machine Learning},
  pages={10842--10851},
  year={2020},
  organization={PMLR}
}

@inproceedings{chen2022hallucinated,
  title={Hallucinated neural radiance fields in the wild},
  author={Chen, Xingyu and Zhang, Qi and Li, Xiaoyu and Chen, Yue and Feng, Ying and Wang, Xuan and Wang, Jue},
  booktitle={Proceedings of the IEEE/CVF Conference on Computer Vision and Pattern Recognition},
  pages={12943--12952},
  year={2022}
}

@inproceedings{ghorbani2019data,
  title={Data shapley: Equitable valuation of data for machine learning},
  author={Ghorbani, Amirata and Zou, James},
  booktitle={International conference on machine learning},
  pages={2242--2251},
  year={2019},
  organization={PMLR}
}

@article{talebi2018nima,
  title={NIMA: Neural image assessment},
  author={Talebi, Hossein and Milanfar, Peyman},
  journal={IEEE transactions on image processing},
  volume={27},
  number={8},
  pages={3998--4011},
  year={2018},
  publisher={IEEE}
}

@article{guo2020fastif,
  title={Fastif: Scalable influence functions for efficient model interpretation and debugging},
  author={Guo, Han and Rajani, Nazneen Fatema and Hase, Peter and Bansal, Mohit and Xiong, Caiming},
  journal={arXiv preprint arXiv:2012.15781},
  year={2020}
}

@inproceedings{martin2021nerf,
  title={Nerf in the wild: Neural radiance fields for unconstrained photo collections},
  author={Martin-Brualla, Ricardo and Radwan, Noha and Sajjadi, Mehdi SM and Barron, Jonathan T and Dosovitskiy, Alexey and Duckworth, Daniel},
  booktitle={Proceedings of the IEEE/CVF conference on computer vision and pattern recognition},
  pages={7210--7219},
  year={2021}
}

@article{kerbl20233d,
  title={3d gaussian splatting for real-time radiance field rendering.},
  author={Kerbl, Bernhard and Kopanas, Georgios and Leimk{\"u}hler, Thomas and Drettakis, George},
  journal={ACM Trans. Graph.},
  volume={42},
  number={4},
  pages={139--1},
  year={2023}
}

@article{jin2021image,
  title={Image matching across wide baselines: From paper to practice},
  author={Jin, Yuhe and Mishkin, Dmytro and Mishchuk, Anastasiia and Matas, Jiri and Fua, Pascal and Yi, Kwang Moo and Trulls, Eduard},
  journal={International Journal of Computer Vision},
  volume={129},
  number={2},
  pages={517--547},
  year={2021},
  publisher={Springer}
}

@inproceedings{helmy2024navigating,
  title={Navigating the World with an Intelligent Tourist Guide Using Generative AI},
  author={Helmy, Mona and El-Din, Yomna Safaa and Mohamed, Omar Tarek and Kader, Omar Salah Abdel and Ramadan, Shehab Adel and Kamal, Amr Essam and Selim, Mohamed Reda Mohamed},
  booktitle={2024 International Telecommunications Conference (ITC-Egypt)},
  pages={1--6},
  year={2024},
  organization={IEEE}
}

@article{mazzacca2023nerf,
  title={NeRF for heritage 3D reconstruction},
  author={Mazzacca, G and Karami, A and Rigon, S and Farella, EM and Trybala, P and Remondino, F and others},
  journal={International Archives of the Photogrammetry, Remote Sensing and Spatial Information Sciences},
  volume={48},
  number={M-2-2023},
  pages={1051--1058},
  year={2023},
  publisher={International Society for Photogrammetry and Remote Sensing}
}

@inproceedings{molaei2023implicit,
  title={Implicit neural representation in medical imaging: A comparative survey},
  author={Molaei, Amirali and Aminimehr, Amirhossein and Tavakoli, Armin and Kazerouni, Amirhossein and Azad, Bobby and Azad, Reza and Merhof, Dorit},
  booktitle={Proceedings of the IEEE/CVF International Conference on Computer Vision},
  pages={2381--2391},
  year={2023}
}

@inproceedings{ren2024nerf,
  title={Nerf on-the-go: Exploiting uncertainty for distractor-free nerfs in the wild},
  author={Ren, Weining and Zhu, Zihan and Sun, Boyang and Chen, Jiaqi and Pollefeys, Marc and Peng, Songyou},
  booktitle={Proceedings of the IEEE/CVF Conference on Computer Vision and Pattern Recognition},
  pages={8931--8940},
  year={2024}
}

@article{oquab2023dinov2,
  title={Dinov2: Learning robust visual features without supervision},
  author={Oquab, Maxime and Darcet, Timoth{\'e}e and Moutakanni, Th{\'e}o and Vo, Huy and Szafraniec, Marc and Khalidov, Vasil and Fernandez, Pierre and Haziza, Daniel and Massa, Francisco and El-Nouby, Alaaeldin and others},
  journal={arXiv preprint arXiv:2304.07193},
  year={2023}
}

@article{kulhanek2024wildgaussians,
  title={{W}ild{G}aussians: {3D} Gaussian Splatting in the Wild},
  author={Kulhanek, Jonas and Peng, Songyou and Kukelova, Zuzana and Pollefeys, Marc and Sattler, Torsten},
  journal={NeurIPS},
  year={2024}
}

@inproceedings{chenrethinking,
  title={Rethinking Atrous Convolution for Semantic Segmentation},
  author={Chen, Liang-Chieh and Papandreou, George and Schroff, Florian and Adam, Hartwig},
  booktitle={Proceedings of the IEEE Conference on Computer Vision and Pattern Recognition (CVPR)},
  year={2017}
}
\end{document}